\documentclass[a4paper,table]{extarticle}
\usepackage{booktabs}
\usepackage{colortbl} 
\usepackage{xcolor}   
\usepackage[table]{xcolor}
\definecolor{pelicanmain}{HTML}{E2E8F0}  
\definecolor{pelicanlight}{HTML}{F8F9FA} 
\usepackage{xspace}
\usepackage{tabularx} 
\usepackage{booktabs} 
\usepackage{graphicx}
\usepackage{natbib}
\usepackage{amssymb}
\usepackage{arpafvg}

\usepackage{multirow}  
\usepackage{colortbl}

\usepackage{helvet}

\usepackage{helvet}
\usepackage{titlesec}
\usepackage{xcolor}
\usepackage{CJKutf8}

\definecolor{titleorange}{RGB}{236,120,47}
\definecolor{rowgreen}{RGB}{234,243,222}
\definecolor{rowblue}{RGB}{230,241,251}
\definecolor{rowpurple}{RGB}{238,237,254}

\usepackage{helvet}
\usepackage{titlesec}
\usepackage{xcolor}

\definecolor{titleorange}{RGB}{236,120,47}

\titleformat{\section}
{\normalfont\fontfamily{phv}\fontseries{bx}\selectfont\Large\color{titleorange}}
{\thesection.}{0.5em}{}

\titleformat{\subsection}
{\normalfont\fontfamily{phv}\fontseries{bx}\selectfont\normalsize\color{black}}
{\thesubsection.}{0.5em}{}

\titleformat{\subsubsection}
{\normalfont\fontfamily{phv}\fontseries{bx}\selectfont\normalsize\color{black}}
{\thesubsubsection.}{0.5em}{}

\usepackage[most]{tcolorbox}
\definecolor{mylightgreen}{HTML}{D7FFD7}
\definecolor{myred}{RGB}{255,0,0}
\definecolor{modiblue}{RGB}{0,40,161}
\usepackage{lipsum}      
\usepackage{booktabs} 
\usepackage{multirow} 
\usepackage{caption} 
\usepackage{makecell}
\usepackage{adjustbox} 
\newtcolorbox{takeawaybox}{
  colback=gray!10,  
  colframe=gray!75, 
  boxrule=0.5pt,    
  arc=2mm,          
  halign=center,    
  valign=center,    
  boxsep=3pt,       
  fontupper=\small\bfseries\sffamily, 
  sharp corners,    
}
\newtcolorbox{casebox}[1][]{
    enhanced,
    colback=black!5!white, 
    colframe=black!60!white, 
    boxrule=0.8pt,           
    arc=2mm,                 
    fonttitle=\bfseries\sffamily, 
    title=#1,                
    left=6pt, right=6pt, top=6pt, bottom=6pt, 
    sharp corners,
    breakable,               
    #1 
}

\usepackage{float} 
\usepackage{epigraph}
\begin{document}
\pagecolor{white}

\makeatletter

\title{\textcolor{titlecolor}{Pelican-VLA 0.5: \\ Attending Before Acting Benefits Generalization}}


\makeatother

\author{
	Beijing Innovation Center of Humanoid Robotics (X-Humanoid) \\
    \textbf{WFM System Group} \\
    \{vito.dai,jian.tang,jason.ju\}@x-humanoid.com \\
    \href{https://github.com/Open-X-Humanoid/Pelican-VLA05}
    {\textcolor{titlecolor}{\texttt{https://github.com/Open-X-Humanoid/Pelican-VLA05}}}\\
    \href{https://huggingface.co/X-Humanoid/Pelican-VLA05}
    {\textcolor{titlecolor}{\texttt{https://huggingface.co/X-Humanoid/Pelican-VLA05}}}
    }

\maketitle

\begin{figure}[h]
    \centering
    \includegraphics[width=\textwidth]{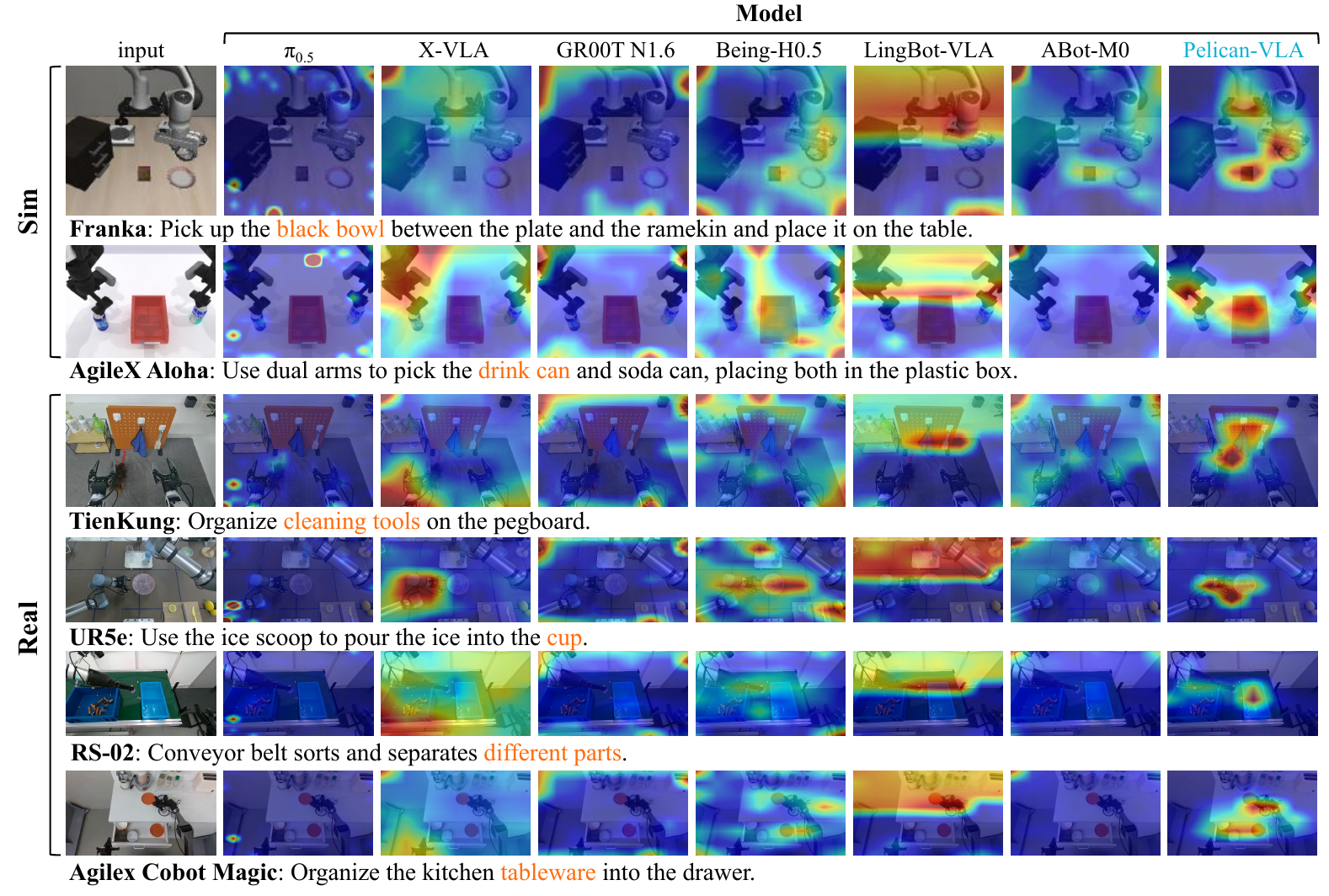}
    \caption{\textbf{Attention visualization comparison with open-source VLA baselines in the zero-shot setting.} Before task-specific fine-tuning, Pelican-VLA 0.5 directs its action-pathway attention to the manipulation-relevant object and contact area. In contrast, other open-source VLA models show more diffuse attention, often spreading over the robot arm, surrounding objects, or background.}
    \label{fig:opensource-vs-pelican}
\end{figure}

\section*{\textbf{Abstract}}

A central goal of Vision-Language-Action (VLA) research is to build robotic models that can truly generalize across objects, scenes, tasks, and embodiments. However, current VLA models still depend heavily on task- and environment-specific robot data, and typically require additional data collection and fine-tuning when transferred to new objects, scenes, tasks, or embodiments. Recent studies suggest that VLA generalization depends not merely on fitting low-level action trajectories, but also on whether task-relevant information can be represented in a transferable form and effectively routed to the action-generation pathway. This suggests that helping the action pathway attend more consistently to manipulation-relevant regions may be beneficial for generalization. Motivated by this view, we analyze the attention of VLA models and find that their action pathways attend \emph{diffusely}, spreading over the robot arm, background, and task-irrelevant objects. 

In this report, we present \textbf{Pelican-VLA 0.5}, a unified VLA model that integrates vision-language understanding, future-frame generation, and action prediction within a single architecture. Pelican-VLA 0.5 achieves \emph{attention-level generalization}: without object annotations, segmentation masks, attention supervision, or task-specific fine-tuning, its action pathway already focuses on the manipulation-relevant object and contact region. This behavior persists across unseen scenes and unseen robot embodiments, and is substantially stronger than in other open-source VLA baselines.  We verify that this ability originates from the learnable \textbf{Bot}tleneck \textbf{Tokens} (BotTokens) inserted between perception and action: by routing task-relevant visual information through a compact bottleneck, the tokens interface induces manipulation-centric attention during pre-training and remains effective across different policy structures, including a MoT-style architecture.

After fine-tuning on RoboTwin, Pelican-VLA 0.5 achieves \textbf{91.4\%} success on \emph{RoboTwin Clean} and \textbf{91.0\%} on \emph{RoboTwin Randomized}, the best average among open-source VLA baselines.  In zero-shot settings, including unseen scenes, unseen objects, and new robot embodiments, Pelican-VLA 0.5 attains non-zero success on several tasks, an early glimmer of generalization. Notably, the attention patterns before and after fine-tuning remain highly similar, suggesting that fine-tuning mainly strengthens the mapping from these pre-formed, manipulation-centric attention regions to executable actions, rather than creating them from scratch.

These findings clarify the meaning of \textbf{Pelican-VLA 0.5}. The model has already achieved strong attention-level generalization during pre-training, but an attention-to-action gap still remains. It can begin to identify \emph{what} to attend to and represent, while reliable execution across new scenes, objects, and embodiments still requires stronger action-level generalization. Pelican-VLA 0.5 is an intermediate model toward truly generalizable VLA models.

\section{Introduction}

Building robots that can follow free-form language instructions and manipulate objects in open, unstructured environments is a central goal of embodied intelligence. Vision-Language-Action (VLA) models have become a leading paradigm toward this goal~\cite{rt2_2023,openvla_2024,octo_2024,pi0_2026,rdt1b_2024,spatialvla_2025,gr00tn1_2025}. By learning an end-to-end mapping from visual observations and language instructions to robot actions, and by building on large-scale multimodal pre-training~\cite{openx_2024,droid_2024,pi05_2025,lap_2026}, these models inherit broad semantic priors and increasingly demonstrate strong performance on everyday manipulation tasks. However, the defining challenge for VLA models is not solving a single benchmark after sufficient adaptation, but \emph{generalization}: a useful robot policy should behave sensibly when the target object, scene, task, or even robot embodiment differs from those seen during training. Despite rapid progress, reliable zero-shot manipulation in unfamiliar settings remains difficult~\cite{openvla_2024,pi05_2025,dreamzero_2026,lap_2026}. 

Recent studies~\cite{pi05_2025,lap_2026,goalvla_2026} suggest that VLA generalization depends not only on fitting low-level action trajectories, but also on how task-relevant information is represented and routed before action prediction. Motivated by this perspective, we analyze the attention maps of representative VLA models. We find that their action pathways often attend \emph{diffusely} over the visual input, with attention spread across the robot arm, background clutter, and task-irrelevant objects, rather than consistently concentrating on the manipulation-relevant target region or potential contact area. One possible reason is that action decoders in VLA models are not explicitly constrained to extract task-critical visual information, and may therefore rely on visual shortcuts or environment-specific correlations~\cite{guidedvla_2026}. This observation motivates us to ask whether VLA models can already form manipulation-relevant attentional representations during pre-training, and whether such representations can provide an early basis for stronger generalization across new scenes, objects, and embodiments.

In this report, we study this question through \textbf{Pelican-VLA 0.5}, a unified VLA model that integrates vision-language understanding, future generation, and action prediction within a single Transformer backbone. Pelican-VLA 0.5 exhibits strong \emph{attention-level generalization}: before any task-specific finetuning, its action pathway already concentrates on manipulation-relevant target regions and contact areas. A central design  is a compact set of learnable BotTokens inserted between perception and action. These tokens serve as a learnable bottleneck between perception and action, querying the upstream vision-language context through full attention and condensing manipulation-relevant information before passing it to the action pathway. Rather than allowing the action pathway to directly attend to dense visual tokens, these BotTokens provide a constrained interface through which task-relevant perceptual information is routed to action generation. These BotTokens encourage the model to organize manipulation-relevant information, including the target object, potential contact region, intended future change, and current robot-state constraint, while reducing direct reliance on low-level visual details. 

In the zero-shot setting, Pelican-VLA 0.5's action pathway already concentrates on the manipulation-relevant object and contact region, as shown in Fig.~\ref{fig:opensource-vs-pelican}. This behavior emerges without object-level supervision and persists across unseen scenes and unseen robot embodiments. Compared with other open-source VLA baselines, Pelican-VLA 0.5 exhibits substantially more target-aligned attention.  Moreover, removing the BotTokens after training does not fully eliminate the model’s ability to attend to the manipulation-relevant object, indicating that the BotTokens helps internalize manipulation-centric attention into the shared backbone and allows the learned representation to persist as a property of the trained model itself.

In RoboTwin simulation, Pelican-VLA 0.5 can perform zero-shot attempts in unseen scenes and new embodiments, as shown in Fig.~\ref{fig:zeroshot}. We fine-tune Pelican-VLA 0.5 on RoboTwin. After fine-tuning, the model achieves strong average success among open-source VLA models, reaching \textbf{91.4\%} on \emph{RoboTwin Clean} and \textbf{91.0\%} on \emph{RoboTwin Randomized}. These results show that the pre-formed manipulation-centric attention can be effectively converted into task success once sufficient action supervision is provided. More importantly, when we visualize the attention maps before and after fine-tuning, we find that the attention patterns remain highly similar, as shown in Fig.~\ref{fig:zeroshot_vs_finetune}. This suggests that fine-tuning Pelican-VLA 0.5 does not primarily create manipulation-centric attention from scratch. Instead, it mainly improves the mapping from manipulation-centric attentions to executable actions.

These findings clarify the meaning of \textbf{Pelican-VLA 0.5}. The model is not presented as a fully zero-shot VLA. Rather, it represents an intermediate stage toward truly generalizable robot policies. It has achieved strong representation-level generalization during pre-training, but a representation-to-action gap still remains. This limitation is partly due to the scale and form of the current training data: Pelican-VLA 0.5 is trained on 2,400 hours of heterogeneous robot data and uses joint-position actions, which are more embodiment-specific than end-effector action representations. Scaling robot data, improving action representations, and strengthening the representation-to-action mapping are therefore necessary next steps toward practical zero-shot manipulation.

\paragraph{Contributions.}
\begin{itemize}
    \item  We introduce \textbf{Pelican-VLA 0.5}, a unified VLA architecture that integrates vision-language understanding, future generation, and action prediction within a shared Transformer backbone. The model uses learnable BotTokens as an intermediate interface between perception and action, providing a compact and extensible architecture.

    \item We identify a form of \textbf{attention-level generalization} that emerges during pre-training. Without task-specific fine-tuning, object annotations, segmentation masks, or attention supervision, Pelican-VLA 0.5 already attends to manipulation-relevant target regions and contact areas across unseen scenes and unseen robot embodiments.
    
    \item We verify that ability stems from the learnable \textbf{BotTokens} inserted between perception and action. These tokens induce manipulation-centric attention patterns by routing task-relevant visual information through a compact perception-to-action interface, and remain effective across different policy structures, including a MoT-style architecture.
    
    \item We clarify the gap between zero-shot representation and zero-shot control. Pelican-VLA 0.5 shows partial zero-shot manipulation ability, and after RoboTwin fine-tuning achieves \textbf{91.4\%} success on \emph{RoboTwin Clean} and \textbf{91.0\%} on \emph{RoboTwin Randomized}. We will release the code, weights, and attention-visualization tools to support reproducibility.
\end{itemize}

\section{Pelican-VLA 0.5}
\label{sec:method}
\subsection{Overview}
\label{sec:method:overview}
\begin{figure}[t]
  \centering
  \includegraphics[width=\linewidth]{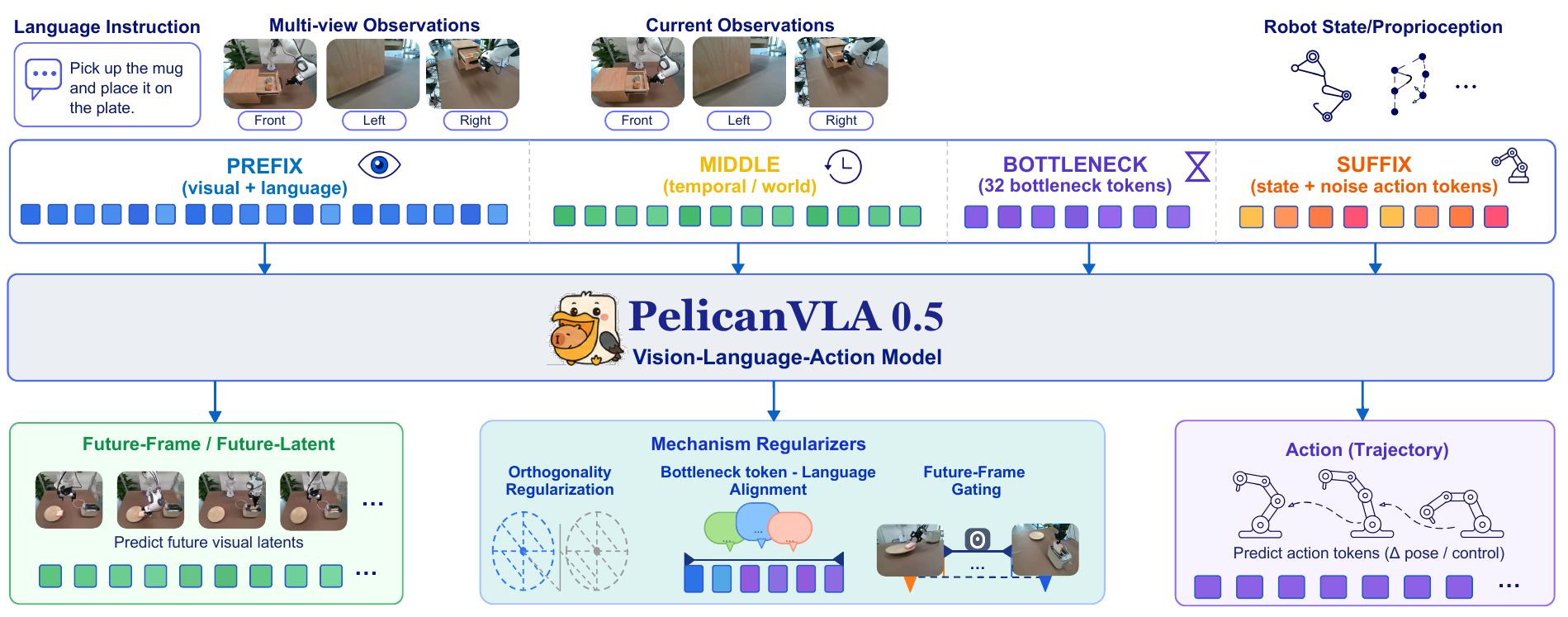}
  \caption{\textbf{Overview of Pelican-VLA 0.5.} Pelican-VLA 0.5 uses a shared Qwen3-VL 4B backbone to unify visual-language understanding, future-frame prediction, and action generation within a single Transformer. The input sequence contains four segments: vision-language \emph{prefix} tokens, Cosmos latent \emph{middle} tokens, $K$ learnable \emph{BotTokens}, and a \emph{suffix} containing proprioceptive states and noisy action chunks. A single forward pass predicts future-frame latents from the middle, a compact task representation from the BotTokens, and action denoising velocities from the suffix. The key design is a bottleneck attention mask: the action pathway reads perception through the BotTokens rather than directly over dense visual tokens. This encourages manipulation-centric attention and allows the denoising loop to reuse a compact BotTokens cache during inference.}
  \label{fig:structure}
\end{figure}

As shown in Fig.~\ref{fig:structure}, Pelican-VLA 0.5 is a unified Vision-Language-Action model built on a single shared Qwen3-VL 4B backbone~\cite{qwen3-vl_2025}. The model is designed to support three tightly coupled functions within one Transformer: visual-language understanding, future-frame prediction, and action generation. Unlike dual-system or Mixture-of-Transformers designs~\cite{internvla-a1_2026,last0_2026}, Pelican-VLA 0.5 does not introduce separate reasoning and acting experts. Instead, all functions operate over a shared token stream and communicate through a common hidden state.

The input sequence is organized into four contiguous segments:
\begin{equation}
  \mathbf Z =
  \big[
  \underbrace{\text{Prefix}}_{\text{vision-language}};
  \underbrace{\text{Middle}}_{\text{Cosmos latents}};
  \underbrace{\text{Bottleneck }}_{K\ \text{bottleneck tokens}};
  \underbrace{\text{Suffix}}_{\text{state and noisy action}}
  \big].
\end{equation}
A single forward pass produces three outputs in parallel: action denoising velocities from the suffix, future-frame latents from the middle, and a compact task representation from the BotTokens.

This section first describes the unified token sequence and attention geometry in \S\ref{sec:method:seq}. We then introduce the BotTokens and its auxiliary regularizers in \S\ref{sec:method:slots}, followed by the training objectives in \S\ref{sec:method:obj} and the cached inference procedure in \S\ref{sec:method:infer}.

\subsection{Unified token sequence}
\label{sec:method:seq}

\paragraph{Prefix: vision-language tokens.}
The current camera observations and language instruction are encoded by the Qwen3-VL backbone. Each camera frame is first processed by the Qwen3-VL vision encoder, and the resulting visual patch embeddings are inserted into the text-token sequence at the corresponding image-token positions. This produces the prefix sequence
\begin{equation}
\mathbf P \in \mathbb{R}^{L_p \times D}.
\end{equation}
The prefix uses bidirectional attention so that language tokens and visual tokens can mutually contextualize each other. This segment provides the semantic visual-language representation of the current scene.

\paragraph{Middle: Cosmos latent tokens.}
In parallel, we encode a short visual history using a frozen Cosmos-Tokenizer~\cite{cosmos-world_2025}. For each view, two frames at time $t{-}15$ and $t$ are mapped into continuous latent features. The features are flattened into the middle sequence
\begin{equation}
\mathbf M \in \mathbb{R}^{L_m \times D}.
\end{equation}
The middle segment provides a pixel-level and dynamics-aware view of the scene, complementing the semantic patch tokens in the prefix. It attends bidirectionally to itself and to the prefix, and is supervised by the future-frame prediction objective described in \S\ref{sec:method:obj}.

\paragraph{Bottleneck: learnable BotTokens}
We introduce $K$ learnable BotTokens,
\begin{equation}
\mathbf S_0 \in \mathbb{R}^{1 \times K \times D},
\end{equation}
initialized from $\mathcal N(0,0.02^2)$ and broadcast across the batch. During training, BotTokens dropout with probability $p=0.1$ is applied:
\begin{equation}
\mathbf S = \mathrm{drop}_p(\mathbf S_0) \in \mathbb{R}^{B \times K \times D}.
\end{equation}
We use $K=32$ by default. Since $K$ is much smaller than the number of upstream visual tokens, the BotTokens form a fixed-capacity interface between perception and action. Their attention to the upstream context provides a direct way to inspect what information is routed into the action pathway.

\paragraph{Suffix: proprioceptive state and noisy action.}
The suffix contains the robot state and the noisy action chunk used by the flow-matching. The proprioceptive state is projected into a single state token. The noisy action chunk
\begin{equation}
\mathbf x_\tau \in \mathbb{R}^{H \times d_a}
\end{equation}
is embedded by a linear layer, concatenated with a sinusoidal embedding of the flow time $\tau$, and fused by a two-layer MLP into $H$ action tokens. We use $H=50$ and $d_a=32$ in our implementation. The suffix sequence is therefore
\begin{equation}
\mathbf U \in \mathbb{R}^{(1+H)\times D}.
\end{equation}
The state token and action tokens are treated as separate attention blocks: the state token provides the current proprioceptive condition, while the action tokens are decoded under the visible context.

\paragraph{Attention geometry.}
The full sequence is
\begin{equation}
\mathbf Z = [\mathbf P;\mathbf M;\mathbf S;\mathbf U].
\end{equation}
We construct a two-dimensional attention mask from per-token block flags, following the prefix-LM and block-causal construction used in prior VLA architectures~\cite{pi0_2026}. The prefix, middle, and BotTokens segments are internally bidirectional. The suffix uses causal or block-causal attention with respect to its visible context. 

When \verb|enable_bottleneck_tokens=true|, the BotTokens segment is inserted between the upstream perception tokens and the action suffix. Disabling the BotTokens recovers a standard dense-attention VLA in which the action suffix can directly attend to the upstream visual context. We use this dense variant as the primary architectural ablation.

\subsection{Bottleneck tokens}
\label{sec:method:slots}
The BotTokens are designed not only as an information bottleneck, but also as an implicit task-reasoning workspace. The BotTokens become an effective bottleneck only when the action suffix cannot bypass them. Pelican-VLA 0.5 enforces this bottleneck through three mechanisms: a curriculum attention mask, an orthogonality regularizer, and a token-gated generation.

\paragraph{Curriculum bottleneck mask.}
Let $\mathcal C = \{\text{prefix}\}\cup\{\text{middle}\}$ denote the upstream perception tokens, and let $\mathcal S$ denote the BotTokens. We define a binary attention-visibility mask $A$, where $A_{i\to j}=1$ means that query token $i$ is allowed to attend to key/value token $j$, and $A_{i\to j}=0$ means that this attention edge is masked out. In the hard bottleneck regime, we remove all direct attention edges from suffix tokens to the upstream perception tokens:
\begin{equation}
A_{i\to j} = 0,
\qquad
\forall i\in\text{suffix},\; j\in\mathcal C.
\end{equation}
The suffix can still attend to the BotTokens and to its permitted suffix context. Thus, the only path from perception to action passes through $\mathcal S$.

Applying this hard constraint from the beginning of training can make optimization unstable. We therefore use a curriculum schedule. At training step $t$, the hard bottleneck mask is applied with probability
\begin{equation}
p_t = \min\left(1,\frac{t}{T_{\mathrm{warm}}}\right),
\qquad T_{\mathrm{warm}}=10\mathrm{k}.
\end{equation}
With probability $1-p_t$, the suffix is allowed to directly attend to the upstream context. This schedule lets the model first learn with an easier dense channel and then progressively shifts perception-to-action communication through the BotTokens.

\paragraph{Orthogonality regularizer.}
A compact tokens set is only useful if different BotTokens capture complementary information. Let $\mathbf Z^{\mathrm{tokens}} \in \mathbb{R}^{B\times K\times D}$ be the final BotTokens outputs of the Transformer. For each batch element $b$, let $\hat{\mathbf Z}^{\mathrm{tokens}}_b\in\mathbb R^{K\times D}$ denote the BotTokens-output matrix after $\ell_2$-normalizing each BotTokens vector along the hidden dimension. We encourage BotTokens diversity by penalizing deviations of the BotTokens Gram matrix from identity:
\begin{equation}
  \mathcal{L}_{\mathrm{reg}}
  =
  \frac{1}{B}\sum_{b=1}^{B}
  \frac{1}{K^2}
  \big\lVert
  \hat{\mathbf Z}^{\mathrm{tokens}}_b
  \big(\hat{\mathbf Z}^{\mathrm{tokens}}_b\big)^\top
  - \mathbf I_K
  \big\rVert_F^2 .
\end{equation}
Here $\mathbf I_K$ is the $K\times K$ identity matrix and $\|\cdot\|_F$ denotes the Frobenius norm. This prevents all BotTokens from collapsing to the same direction and encourages them to encode distinct factors of the scene, task, or action context.

\paragraph{Bottleneck tokens gating of generation.}
The BotTokens are also connected to the future-frame prediction branch. From the mean-pooled BotTokens output
\begin{equation}
\bar{\mathbf z}
=
\frac{1}{K}
\sum_{k=1}^{K}\mathbf z^{\mathrm{tokens}}_k,
\end{equation}
we compute a channel-wise gate $\mathbf g=\sigma(\mathbf W_g \bar{\mathbf z})$, where $\mathbf W_g$ projects the pooled BotTokens feature to the channel dimension of the generation features. The gate modulates the generation features before they are decoded back into Cosmos latent space:
\begin{equation}
\tilde{\mathbf F}_{\mathrm{gen}}
=
\mathbf F_{\mathrm{gen}} \odot \mathbf g.
\end{equation}
Here $\mathbf g$ is broadcast over the token or spatial dimensions of $\mathbf F_{\mathrm{gen}}$. This encourages the BotTokens to contain information that is useful not only for action prediction, but also for predicting what will happen next. In this way, the BotTokens representation is tied to forward-predictive, action-relevant structure rather than static appearance alone.

\subsection{Training objectives}
\label{sec:method:obj}

Pelican-VLA 0.5 is trained with a multi-task objective that combines action generation, future-frame prediction, language-task alignment, and BotTokens regularization.

\paragraph{Flow-matching action loss.}
We use conditional flow matching~\cite{flow_2023,pi0_2026} to model the action chunk. Let $\mathbf a \in \mathbb{R}^{H\times d_a}$ be the target delta-action chunk, where $H$ is the action horizon and $d_a$ is the action dimension. We sample Gaussian noise $\boldsymbol\varepsilon \sim \mathcal N(\mathbf 0,\mathbf I)$ with the same shape as $\mathbf a$, and sample the flow time $\tau\sim\mathcal U(0,1)$. We construct the interpolated noisy action and the target velocity as
\begin{equation}
  \mathbf x_\tau = \tau\,\boldsymbol\varepsilon + (1-\tau)\,\mathbf a,
  \qquad
  \mathbf u = \boldsymbol\varepsilon - \mathbf a .
\end{equation}
Here $\mathbf x_\tau$ moves from the clean action $\mathbf a$ at $\tau=0$ to Gaussian noise $\boldsymbol\varepsilon$ at $\tau=1$, and $\mathbf u$ is the constant velocity along this linear path. The noisy action $\mathbf x_\tau$ is embedded into the suffix, and the action-token outputs are passed through a LayerNorm-MLP head $v_\theta$. The action loss is
\begin{equation}
  \mathcal{L}_{\mathrm{action}}
  = \mathbb{E}_{\mathbf a,\boldsymbol\varepsilon,\tau}
    \big\lVert v_\theta(\mathbf Z)_{\mathrm{act}} - \mathbf u \big\rVert_2^2 .
\end{equation}
Here $v_\theta(\mathbf Z)_{\mathrm{act}}\in\mathbb R^{H\times d_a}$ denotes the decoded outputs corresponding only to the $H$ action tokens.

\paragraph{Future-frame generation loss.}
The middle outputs are decoded by the BotTokens-gated generation head and supervised in the frozen Cosmos latent space. Given the future frame $\mathbf I_{t+15}$, we first encode it into a target Cosmos latent $\mathbf c_{t+15}=\mathrm{Cosmos}(\mathbf I_{t+15})$. We compute
\begin{equation}
  \mathcal{L}_{\mathrm{gen}}
  = \big\lVert \mathrm{Dec}\big(\tilde{\mathbf F}_{\mathrm{gen}}\big)
              - \mathbf c_{t+15}\big\rVert_2^2 .
\end{equation}
Here $\mathrm{Dec}(\cdot)$ denotes the latent prediction head that maps the gated generation features to the Cosmos latent space. The loss is computed over valid cameras only. Predicting in Cosmos latent space provides a compact dynamics signal and avoids the cost and instability of direct pixel reconstruction~\cite{futurevla_2026}.

\paragraph{BotTokens-language contrastive alignment.}
We further align the trajectory representation with the language instruction using a symmetric InfoNCE objective~\cite{infonce_2018,clip_2021,lara_2026}. For a mini-batch of size $B$, let $\mathbf Z_{\mathrm{traj}}\in\mathbb R^{B\times d_c}$ denote the trajectory embeddings obtained by mean-pooling and projecting the selected trajectory feature source, and let $\mathbf Z_{\mathrm{text}}\in\mathbb R^{B\times d_c}$ denote the text embeddings obtained by mean-pooling the instruction token embeddings. Both embeddings are $\ell_2$-normalized along the embedding dimension, where $d_c$ is the contrastive embedding dimension. With learnable temperature $\exp(s)$, the batchwise similarity logits are
\begin{equation}
  \mathbf L = \exp(s)\,\mathbf Z_{\mathrm{traj}}\mathbf Z_{\mathrm{text}}^\top .
\end{equation}
The contrastive loss is
\begin{equation}
  \mathcal{L}_{\mathrm{task}}
  = \tfrac12\big(\mathrm{CE}(\mathbf L,\mathbf y)
                + \mathrm{CE}(\mathbf L^\top,\mathbf y)\big),
  \qquad y_i = i,\; i=1,\ldots,B .
\end{equation}
Here $\mathbf y$ is the identity matching target within the mini-batch: the $i$-th trajectory is paired with the $i$-th instruction. When different samples share the same instruction, the corresponding off-diagonal pairs are masked out of the denominator to avoid treating true semantic matches as negatives.

\paragraph{Total objective.}
The final objective is
\begin{equation}
  \mathcal{L}
  = \mathcal{L}_{\mathrm{action}}
+ \lambda_{\mathrm{gen}}\,\mathcal{L}_{\mathrm{gen}}
+ \lambda_{\mathrm{task}}\,\mathcal{L}_{\mathrm{task}}
+ \lambda_{\mathrm{reg}}\,\mathcal{L}_{\mathrm{reg}} .
\end{equation}
In our pre-training runs, we set
$\lambda_{\mathrm{gen}}=0.01$,
$\lambda_{\mathrm{task}}=0.1$,
$\lambda_{\mathrm{reg}}=0.01$. The bottleneck schedule is advanced by the global training step, so the perception-to-action pathway is gradually constrained as optimization proceeds.

\subsection{Cached inference}
\label{sec:method:infer}

At inference time, the BotTokens provide a compact cacheable interface for action denoising. The prefix and middle tokens depend only on the current visual-language context, not on the flow time. Therefore, they are computed once and cached. Inference proceeds in three stages.

First, the prefix is processed by the language model with key-value caching enabled. Second, the middle tokens and BotTokens extend the cache. After this stage, we keep only the $K$ BotTokens key-value pairs as the visual context for the denoising loop and discard the much larger prefix and middle cache. Third, we initialize the action chunk from Gaussian noise, $\mathbf x_1\sim\mathcal N(\mathbf 0,\mathbf I)$, and integrate the flow ODE backward from $\tau=1$ to $\tau=0$ using $N=10$ Euler steps. At each step, only the suffix tokens are re-embedded and passed through the model using a copy of the BotTokens cache:
\begin{equation}
\mathbf x_{\tau+\Delta \tau}
=
\mathbf x_\tau
+
\Delta \tau \cdot v_\theta(\mathbf Z)_{\mathrm{act}},
\qquad
\Delta \tau=-\frac{1}{N}.
\end{equation}
Here $v_\theta(\mathbf Z)_{\mathrm{act}}\in\mathbb R^{H\times d_a}$ denotes the predicted velocity for the $H$ action tokens. After the final step reaches $\tau=0$, the predicted delta actions are de-normalized and added to the current proprioceptive state. This procedure makes the repeated denoising cost depend on the suffix length and the number of BotTokens, rather than on the full visual context length. As a result, the model can preserve a BotTokens-mediated perception-to-action interface while supporting efficient high-frequency control.

\section{Experiments}
\label{sec:experiments}

\subsection{Training data}
\label{sec:exp:data}
Pelican-VLA 0.5 is pre-trained on a heterogeneous, cross-embodiment mixture of large-scale manipulation corpora, including AgiBot World Alpha~\cite{agibot-world-colosseo_2025}, InternData-A1~\cite{interndata-a1_2025}, the Galaxea Open-World Dataset~\cite{galaxea-open-world-dataset-g0_2025}, and roughly $1000$ hours of self-collected teleoperation Tienkung and UR data whose collection format is comparable to RoboMIND~\cite{robomind_2025}. The pre-training mixture contains more than $6000$ hours of manipulation data. All datasets are unified through the LeRobot data interface. Since the corpora cover different robot embodiments with different degrees of freedom, we pad all state and action vectors to a common $32$-dimensional format. This allows the model to train jointly across heterogeneous embodiments while using a shared action head. Unless otherwise specified, actions are represented in joint-position space.

\subsection{Implementation and pre-training recipe}
\label{sec:method:impl}
We instantiate the backbone as the Qwen3-VL 4B and train it end-to-end in \verb|bfloat16| with gradient checkpointing. The action chunk length is $H{=}50$ with delta-action targets, and the state and action vectors are padded to $32$ dimensions for cross-embodiment training. Each observation comprises three camera views resized to $224{\times}224$, with image deltas of $[-15,0,15]$. Optimization uses AdamW with a peak learning rate of $5{\times}10^{-5}$, $2$k warmup steps, and cosine decay to $5{\times}10^{-6}$, together with a per-GPU batch size of $6$. The reasoning-BotTokens module uses $K{=}32$ BotTokens with dropout $0.1$ and a curriculum warmup of $10$k steps, and the InfoNCE term is enabled with a task-embedding dimension of $256$. The current model is pre-trained for only $0.4$ epoch over this mixture, so it has effectively seen about $2400$ hours of data; the results reported here are therefore obtained under a substantially under-trained regime. 

\subsection{Simulation results}
\label{sec:exp:sim}
On the RoboTwin benchmark~\cite{robotwin_2025}, Pelican-VLA attains an average success rate of $\mathbf{91.4\%}$ under the \emph{clean} setting and $\mathbf{91.0\%}$ under the \emph{randomized} setting, as summarized in Table~\ref{tab:benchmark_results}. These results are obtained with our proposed adaptive action execution strategy, which dynamically adjusts action execution according to the model’s prediction uncertainty. The gap between the two settings is only $0.4$ points, indicating that the policy’s competence does not rely on the nominal appearance of the scene. This is consistent with our hypothesis that routing perception through a compact BotTokens suppresses low-level visual shortcuts and instead encodes task-relevant, manipulation-relevant structure that transfers across heavy randomization.

\definecolor{pelicanrow}{RGB}{232,242,255}
\begin{table}[t]
\small
\centering
\caption{Benchmark results on seen tasks in RoboTwin. Compared with representative recent VLA baselines, Pelican-VLA 0.5 achieves the best performance in both clean and randomized settings, yielding the highest average success rate. Unless otherwise specified, best results are highlighted in \textbf{bold}, and second-best results are \underline{underlined}.}
\label{tab:benchmark_results}
\begin{tabular}{lccc}
\toprule
Methods & Clean & Randomized & Average \\ \midrule
$\pi_{0}$~\cite{pi0_2026}               & 80.0 & 79.5 & 79.8 \\
$\pi_{0.5}$~\cite{pi05_2025}            & 86.8 & 87.0 & 86.9 \\
X-VLA~\cite{x-vla_2025}                 & 72.9 & 72.8 & 72.9 \\
StarVLA-OFT~\cite{starvla_2026}         & 88.2 & 88.3 & 88.3 \\
ABot-M0~\cite{abot-m0_2026}             & 86.1 & 85.1 & 85.6 \\
LingBot-VLA~\cite{lingbot-vla_2026}     & 88.6 & 86.7 & 87.7 \\
Qwen-VLA~\cite{qwen-vla_2026}           & 86.1 & 87.2 & 86.7 \\
JoyAI-RA~\cite{joyai-ra_2026}           & 90.5 & 89.3 & 89.9 \\
Hy-VLA~\cite{hy-embodied-0.5-vla_2026}	&\underline{90.9}	&\underline{90.1} & \underline{90.5}\\
\midrule
\cellcolor{pelicanrow}\textbf{Pelican-VLA 0.5} & \cellcolor{pelicanrow}\textbf{91.4} & \cellcolor{pelicanrow}\textbf{91.0} & \cellcolor{pelicanrow}\textbf{91.2} \\ \bottomrule
\end{tabular}
\vspace{0.2em}
\end{table}

\paragraph{Zero-shot generalization.}

Beyond the in-distribution benchmark above, we probe whether Pelican-VLA can act \emph{without any task-specific fine-tuning}. We deploy the pre-trained policy directly on RoboTwin 2.0, which is entirely held out from pre-training, under conditions it has never seen during training, including novel objects, novel scene layouts, and an \emph{new robot embodiment}. To avoid inflating zero-shot performance through task-level leakage, we first audit the overlap between InternData-A1 and RoboTwin 2.0. We compute TF-IDF cosine similarity between task prompts from InternData-A1 and RoboTwin 2.0, manually re-examine all candidate pairs by reading the task descriptions and, when necessary, the corresponding demonstration videos, and exclude all tasks judged to be highly or moderately suspicious overlaps.

\begin{figure}[t]
  \centering
  \includegraphics[width=0.95\linewidth]{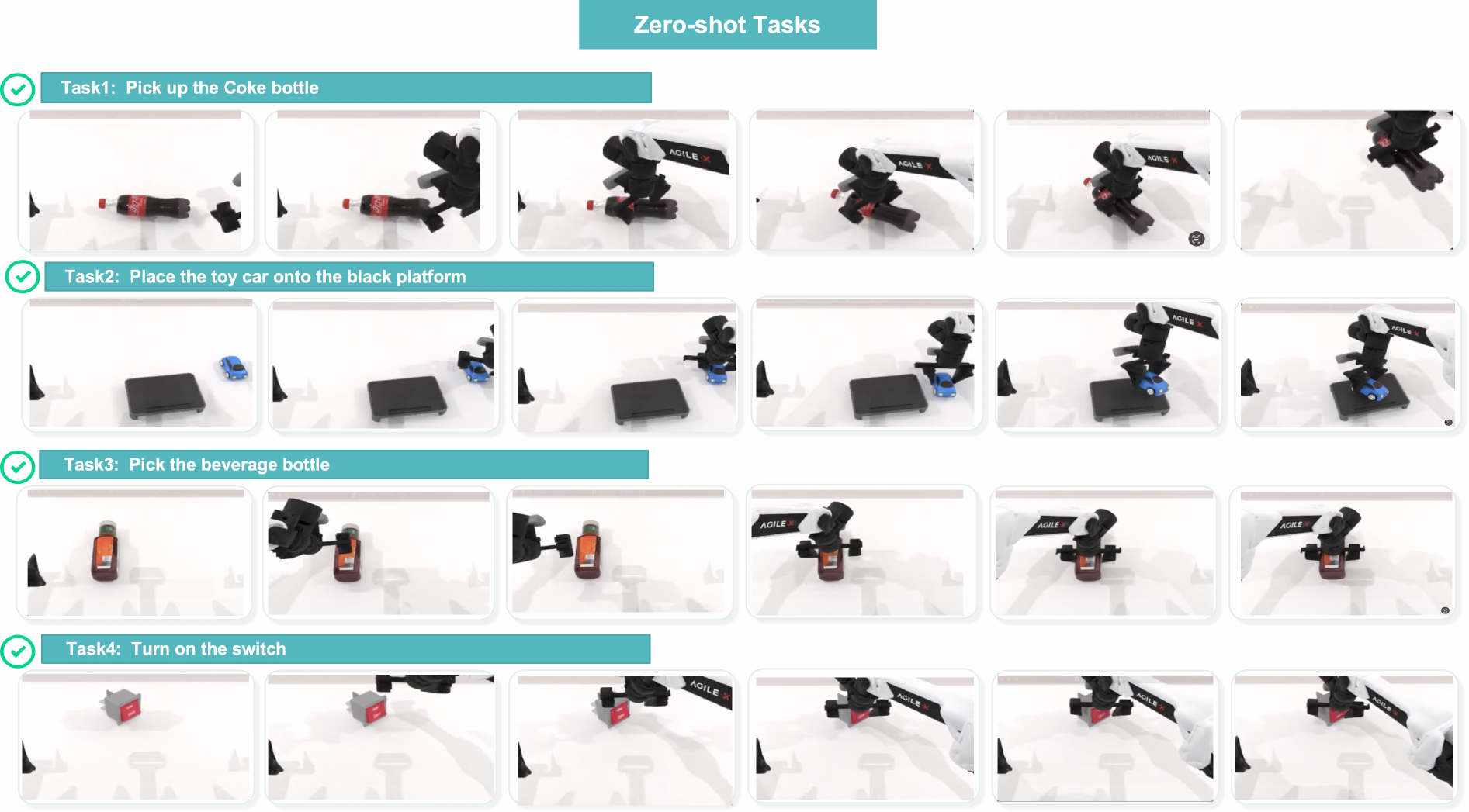}
  \caption{\textbf{Zero-shot rollouts before task-specific fine-tuning.} Pelican-VLA 0.5 is deployed directly on RoboTwin in unseen scenes. The policy often approaches the manipulation-relevant object and produces coherent task-directed motions, including pick up bottle, beverage-bottle grasping, and switch activation.}
  \label{fig:zeroshot}
\end{figure}

As shown in Fig.~\ref{fig:zeroshot}, Pelican-VLA exhibits early zero-shot manipulation behavior on tasks such as picking up bottle, placing a toy car onto a platform, picking up a beverage bottle, and turning on a switch. The policy reaches toward the correct manipulation-relevant object and produces coherent, goal-directed motions rather than random or degenerate behavior, indicating that the manipulation-centric attention formed during pre-training transfers to unseen objects, scenes, and embodiments. However, this behavior remains imperfect: success rates in this strict zero-shot regime are still low, and failures typically occur at fine-grained stages such as stable grasping and precise placement rather than target selection or approach. This exposes the current representation-to-action gap in Pelican-VLA: learning to \emph{see} the target is already emerging, while learning to \emph{act} on it reliably remains the outstanding challenge. We believe this remaining gap is primarily data-driven, arising from the limited scale of the training data and the relatively weak cross-embodiment generalization of joint-space action representations, rather than from the BotTokens itself. As such, we expect the gap to be reduced through scaling robot data and improving action parameterizations, without changing the core BotTokens-mediated architecture.

\subsection{Real-world Robot Evaluation}
\label{sec:exp:real}

\begin{figure}[t]
  \centering
  \includegraphics[width=0.95\linewidth]{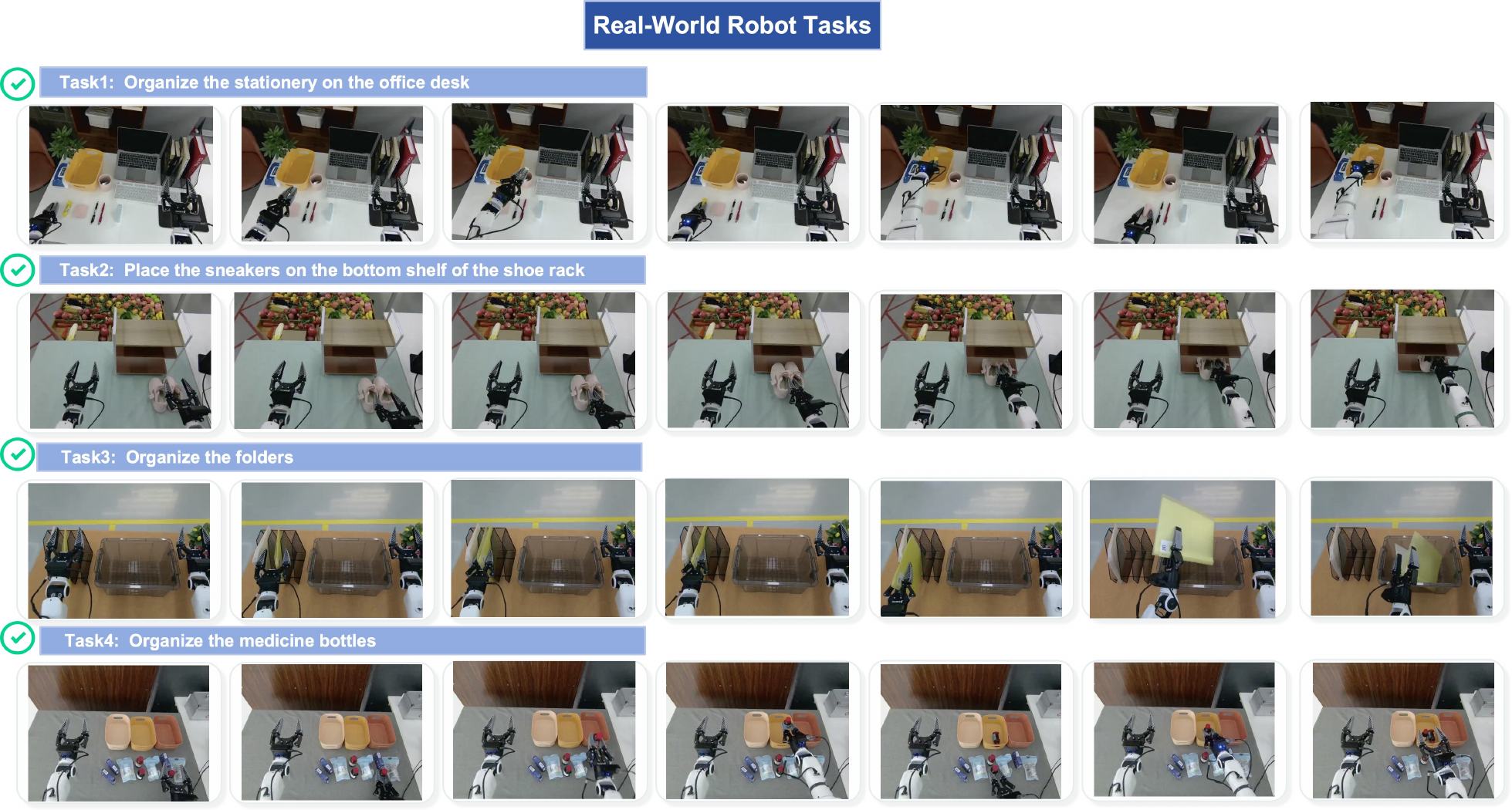}
  \caption{\textbf{Real-world tabletop cleanup on the TienKung humanoid.} Pelican-VLA 0.5 is fine-tuned on teleoperated demonstrations and deployed on a cluttered tabletop-cleanup task. }
  \label{fig:realrobot}
\end{figure}

\paragraph{Tabletop cleanup on TienKung.}
We further evaluate Pelican-VLA 0.5 on a real-world tabletop-cleanup task using the TienKung humanoid~\cite{tienkung}, as shown in Fig.~\ref{fig:realrobot}. In this task, the robot must clear assorted objects from a cluttered desktop and place them into a designated container. The task requires long-horizon manipulation, repeated object selection, stable grasping, and robust execution under real-world sensing and actuation noise.

We fine-tune the pre-trained policy on teleoperated demonstrations collected on the TienKung platform and evaluate it over a fixed set of trials with randomized object types, object counts, and initial placements. This experiment tests whether the pre-trained representation can support downstream adaptation beyond simulation. As shown in Fig.~\ref{fig:realrobot}, Pelican-VLA 0.5 completes the tabletop-cleanup task with a success rate of 80\%, consistently selecting the correct objects, executing stable grasps, and placing them into the target container.

\section{Analysis}
\label{sec:analysis}
In the zero-shot setting shown in Fig.~\ref{fig:opensource-vs-pelican}, without task-specific fine-tuning and without object-, segmentation-, attention-, or reasoning-trace supervision, the attention maps of Pelican-VLA 0.5’s action pathway already highlight the manipulation-relevant object and its contact region. This manipulation-relevant pattern is not directly supervised. It is also not explicitly imposed as an attention target. The goal of this section is therefore to identify where this representation comes from, how it develops during training, and why it does not yet fully translate into reliable zero-shot manipulation.

We combine qualitative attention visualization with an IoU-based grounding analysis. Following prior work on attention visualization in Transformers~\cite{attnrollout_2020}, we visualize where the action pathway attends over visual tokens. This protocol allows us to study whether the model routes perception through manipulation-relevant regions, and whether this routing changes with architecture, training objectives, pre-training progress, and fine-tuning.

\subsection{Disentangling data from architecture}
\label{sec:analysis:arch}
A first hypothesis is that the representation is already latent in the demonstrations, such that \emph{any} sufficiently capable policy trained on them would acquire it; were this the case, the phenomenon would be uninformative about our design. We evaluate this hypothesis under a minimal intervention: holding the data, optimizer, schedule, and parameter budget fixed, we vary only the architecture, training a dense Mixture-of-Transformers (MoT) policy~\cite{internvla-a1_2026,last0_2026} whose action expert attends \emph{densely} to all upstream perception.

\begin{figure}[t]
  \centering
\includegraphics[width=0.8\linewidth]{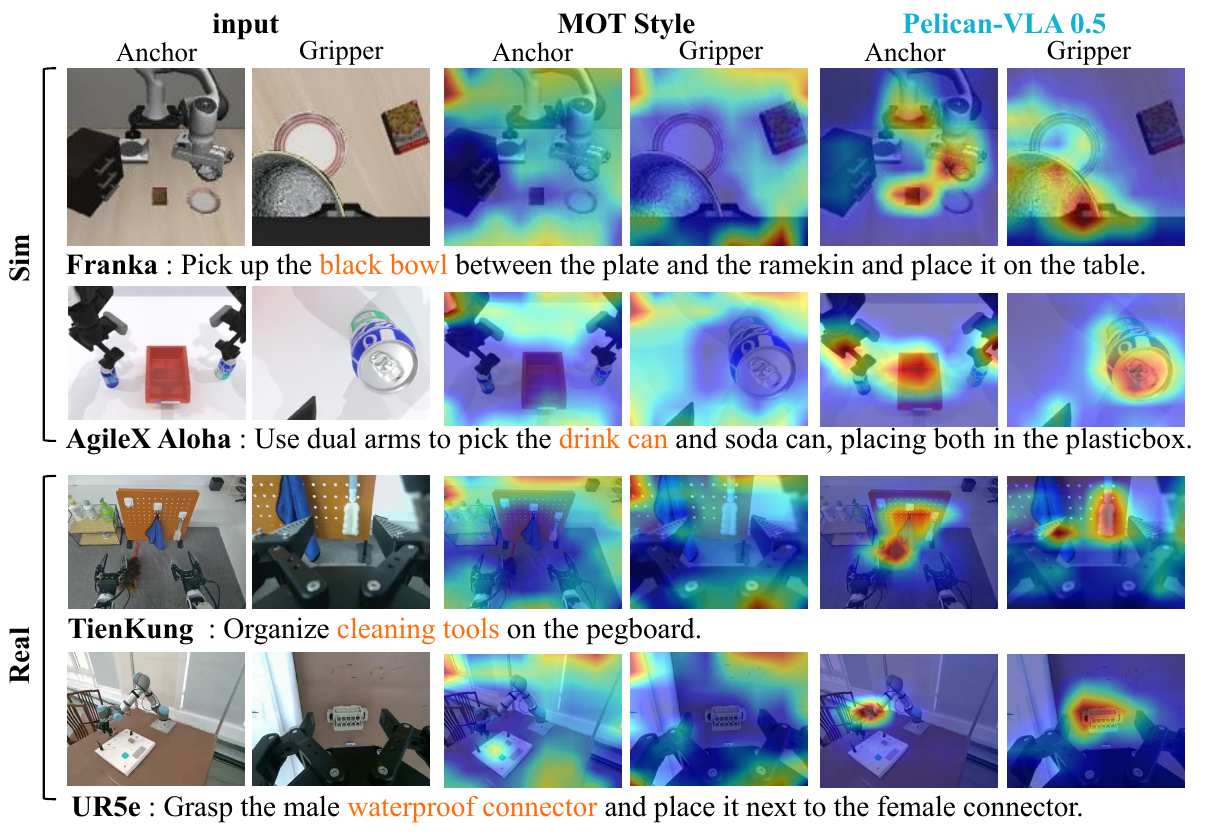}
  \caption{\textbf{Comparison of attention patterns between Pelican-VLA 0.5 and a dense MoT baseline under matched data, optimization, and parameter budgets.} Although both models achieve comparable action supervision objectives, the dense MoT baseline exhibits diffuse attention, whereas Pelican-VLA 0.5 forms concentrated manipulation-relevant attention patterns around manipulation-relevant regions.}
  \label{fig:mot_vs_pelican}
\end{figure}

The two models differ substantially, as shown in Fig.~\ref{fig:mot_vs_pelican}. The dense MoT policy reaches a comparable training action loss, indicating that it can fit the demonstrations under the same data and compute budget. However, its attention remains diffuse and only weakly aligned with the target object. In contrast, Pelican-VLA 0.5 forms a concentrated, manipulation-relevant attention pattern. Since the training data and optimization setup are matched, the difference cannot be explained by the corpus alone. This suggests that the representation is induced by the architecture, and more specifically by how perception is routed to action.

\subsection{Ablation}
\label{sec:analysis:ladder}
Having established that the effect is architectural, we next identify which component is responsible. Pelican-VLA 0.5 differs from a plain action model in several ways: it uses a future-frame generation loss, a BotTokens-language contrastive loss, and the BotTokens bottleneck. The full model alone does not reveal which component causes the observed representation.

We therefore construct an ablation ladder that introduces one mechanism at a time while keeping the backbone, data, and training budget fixed. The ladder contains five configurations:
\begin{enumerate}
\item action loss only;
\item action loss $+$ future-frame generation loss $\mathcal{L}_{\mathrm{gen}}$;
\item action loss $+$ language contrastive loss $\mathcal{L}_{\mathrm{task}}$, applied to a dense model without BotTokens;
\item action loss $+$ the BotTokens.
\item action loss $+$ the BotTokens  $+$ contrastive loss;
\end{enumerate}

\begin{figure}[t]
  \centering
    \includegraphics[width=\linewidth]{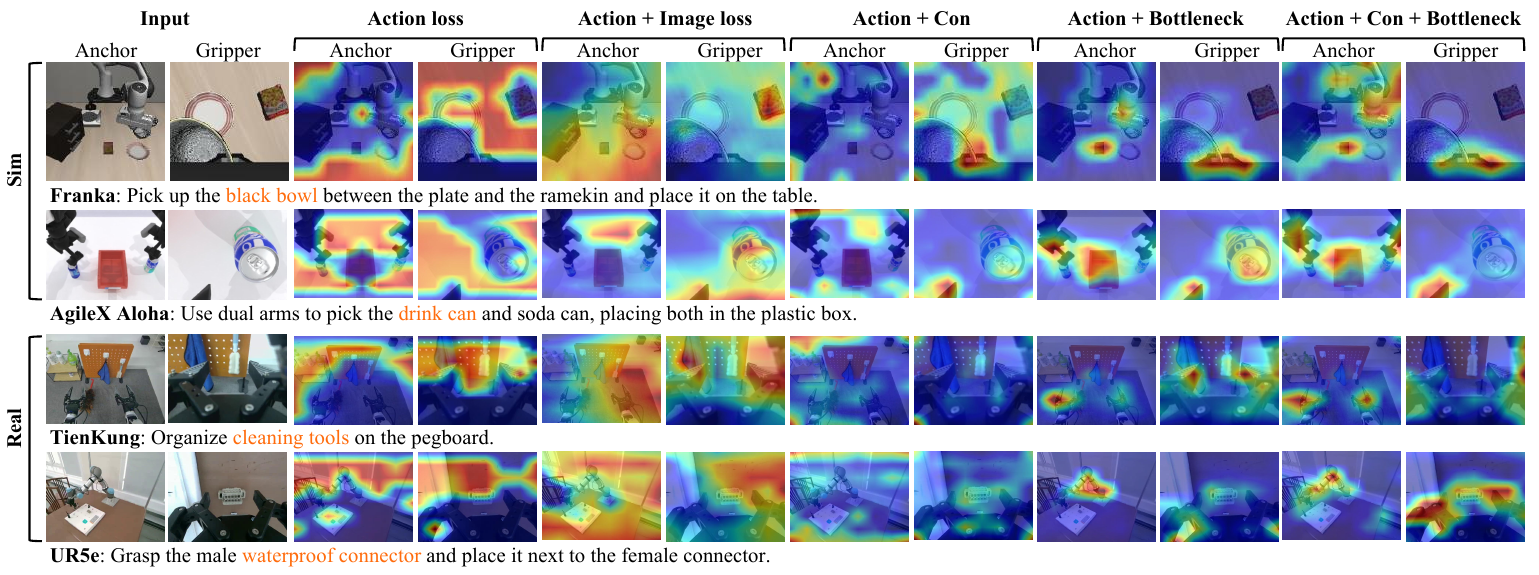}

  \caption{\textbf{Component-wise ablation of Pelican-VLA 0.5.} Adding each training objective individually shows that auxiliary losses alone do not induce manipulation-relevant attention; the manipulation-relevant pattern emerges when perception is routed through the BotTokens.}
  \label{fig:ablation}
\end{figure}

The ablation yields a clear result. The generation loss leaves the representation unchanged, and the contrastive loss imposed on a dense model likewise has no effect---both configurations leave the attention near chance. The representation emerges \emph{only} at row 4, precisely when perception is routed through the BotTokens, at which point it becomes pronounced. That a single mechanism accounts for the entire effect is notable; the evidence nonetheless is unambiguous: the auxiliary losses are not the cause, whereas the BotTokens is. Together with \S\ref{sec:analysis:arch}, this completes the attribution: the representation is induced specifically by routing perception through the narrow BotTokens.

\begin{figure}[t]
  \centering
    \includegraphics[width=0.7\linewidth]{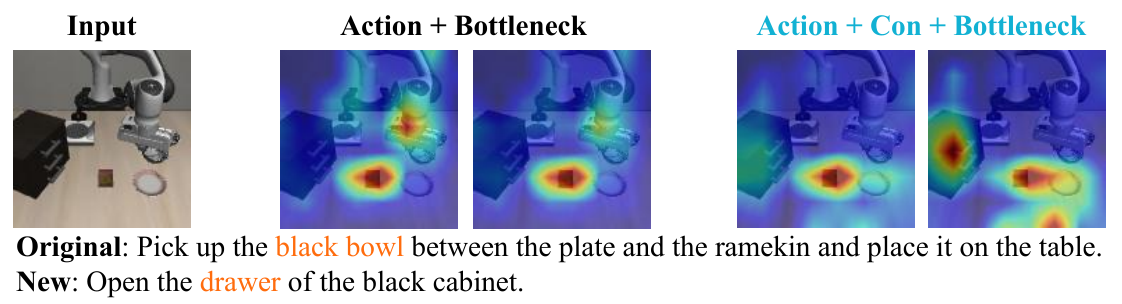}
  \caption{\textbf{Language-conditioned object grounding.} For the same visual scene, changing the instruction shifts the attention toward the newly specified target object. This instruction-dependent behavior becomes stronger with the BotTokens-language contrastive objective, indicating improved alignment between language instructions and manipulation-centric attentions.}
  \label{fig:lang_swap}
\end{figure}

\paragraph{Language-conditioned attention.}
When examining intermediate checkpoints along this training ladder, we observed a phenomenon that cannot be fully explained by the BotTokens alone. In an early checkpoint, as shown in Fig.~\ref{fig:lang_swap}, the BotTokens-only model tends to localize \emph{an} object, often the visually dominant or frequently manipulated one, regardless of the instruction. After introducing the contrastive task loss, the same scene exhibits language-conditioned behavior: changing the instruction from \emph{Pick up the black bowl between the plate and the ramekin and place it on the table.''} to \emph{Open the drawer of the black cabinet.’’} shifts the attention from the bowl to the cabinet, following the object specified by the language. This suggests that the contrastive loss does not create attentional focus by itself, but can make the BotTokens content more conditional on language rather than only on visual salience.

However, this language-conditioned attention pattern appears mainly in early checkpoints and becomes less evident in later stages of training. We hypothesize that this degradation is related to noise in the training data, where a non-negligible portion of language instructions does not precisely match the executed task. As training proceeds, such language-action mismatch may weaken the language-conditioned representation signal. Therefore, we treat this observation as evidence that controllable representation can emerge from the combination of BotTokens and $\mathcal{L}_{\mathrm{task}}$, while also noting that its stability depends on the quality of language-action alignment in the data. In the final full-scale model, we will address this issue through stricter data curation and improved language-action alignment, so that language-conditioned representation can be preserved throughout training.

\begin{figure}[t]
  \centering
    \includegraphics[width=0.6\linewidth]{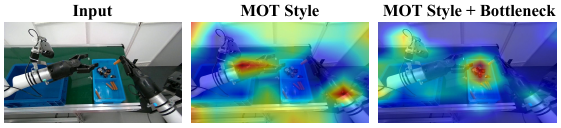}
  \caption{\textbf{Bottleneck tokens transferability across architectures.} We insert the same BotTokens interface into a MoT-style architecture to test whether the BotTokens-induced attention pattern depends on the specific Pelican-VLA design. The resulting model still concentrates its attention on manipulation-relevant object regions, suggesting that BotTokens can serve as a transferable interface for routing task-relevant visual information toward action generation across different VLA architectures.}
  \label{fig:mot_slot}
\end{figure}

\paragraph{Bottleneck tokens transferability across architectures.}
To further examine whether the effect of BotTokens is tied to the specific Pelican-VLA architecture, we also insert the same BotTokens interface into a MoT-style architecture, as illustrated in Fig.~\ref{fig:mot_slot}. The resulting model exhibits a similar attention pattern: despite the change in backbone organization and action-prediction pathway, the BotTokens consistently concentrate on regions related to the manipulated object and its contact area. This suggests that the BotTokens are not merely an artifact of our particular architecture, but can serve as a more general interface for routing task-relevant visual information toward action generation. While the strength and language selectivity of this behavior may still depend on the training objective and data quality, the MoT-based result indicates that BotTokens provide a transferable mechanism for inducing manipulation-centric attention across different VLA designs.

\subsection{Attention dynamics over pre-training}
\label{sec:analysis:emerge}
\begin{figure}[t]
  \centering
 \includegraphics[width=0.9\linewidth]{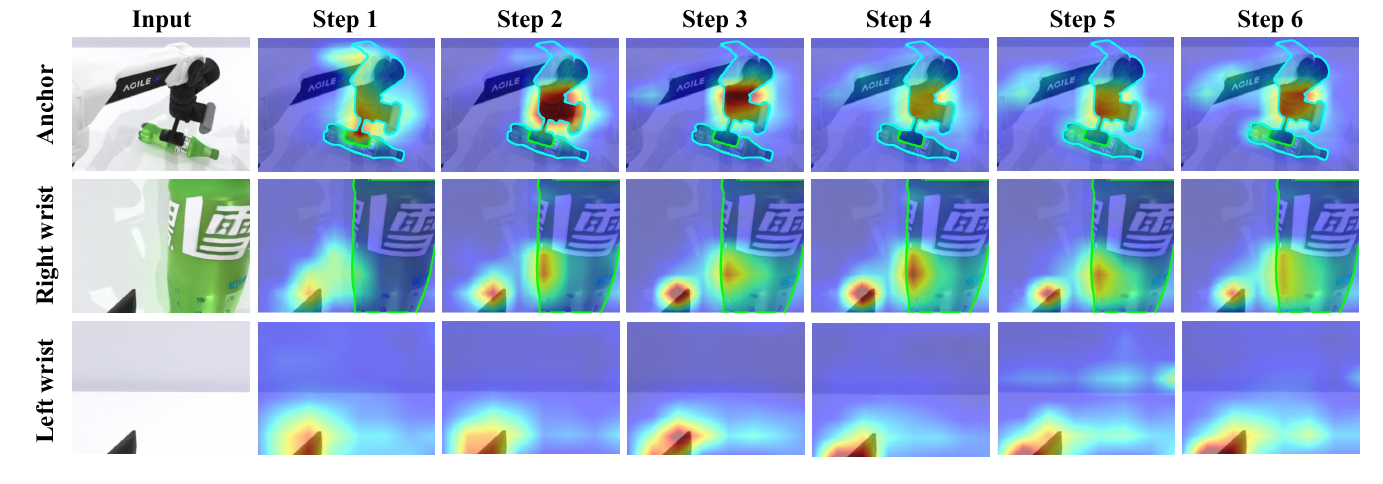}

  \caption{\textbf{The representation builds up over pre-training.} The BotTokens attention starts diffuse and migrates onto the manipulation-relevant object as training proceeds; the representation keeps rising after the action loss has saturated.}
  \label{fig:attn_evolution}
\end{figure}

To characterize how representation emerges during pre-training, we track attention through the BotTokens across checkpoints of the full model, as shown in Fig.\ref{fig:attn_evolution} and Table\ref{tab:attention_spearman}.

As pre-training progresses, attention first becomes spatially concentrated and only later becomes aligned with the object specified by the instruction. This temporal ordering suggests a two-stage process: the model first learns to compress visual information into compact regions, and then learns to select the semantically relevant object among them.

Table~\ref{tab:attention_spearman} provides quantitative evidence for this trend. Manipulator-related regions remain a strong attention recipient throughout pre-training, with an IoU around 0.35 and no significant monotonic trend ($\rho=-0.43$, $p=0.21$). In contrast, attention alignment with the target object increases significantly, with IoU rising from 0.054 to 0.124 and showing a strong positive monotonic trend across checkpoints ($\rho=0.76$, $p=0.01$). Thus, although the absolute attention mass remains partly biased toward the robot embodiment, the systematic change during pre-training is directed toward task-relevant objects.

This object-directed migration continues even after the action loss has largely plateaued, suggesting that representation is not merely a by-product of fitting the action objective. Instead, it reflects a slower reorganization of how task-relevant visual information is compressed through the BotTokens. Since no object-level or attention supervision is provided, this trajectory supports our view that manipulation-centric attention can emerge through pre-training under a BotTokens-mediated architecture.

In addition, removing the BotTokens after training does not fully eliminate the model’s ability to attend to the manipulation-relevant object, indicating that the BotTokens help internalize manipulation-centric attention into the shared backbone and allows the learned representation to persist as a property of the trained model itself.

\begin{table}[t]
\centering
\small
\begin{tabular}{lcccc}
\hline
\textbf{Region} & \textbf{IoU (50k)} & \textbf{IoU (500k)} & \textbf{$\Delta$IoU} & \textbf{Spearman $\rho$} \\
\hline
Manipulator (arm + gripper) & 0.360 & 0.350 & -0.010 & -0.43 (p = 0.21) \\
Target object (bottle) & 0.054 & 0.124 & +0.070 & 0.76 (p = 0.01) \\
\hline
\end{tabular}
\caption{\textbf{Attention dynamics across pre-training checkpoints.} While manipulator attention remains stable and dominant, target object attention shows a strong monotonic increase, as evidenced by a significant positive Spearman correlation.}
\label{tab:attention_spearman}
\end{table}

\begin{figure}[t]
  \centering

    \includegraphics[width=\linewidth]{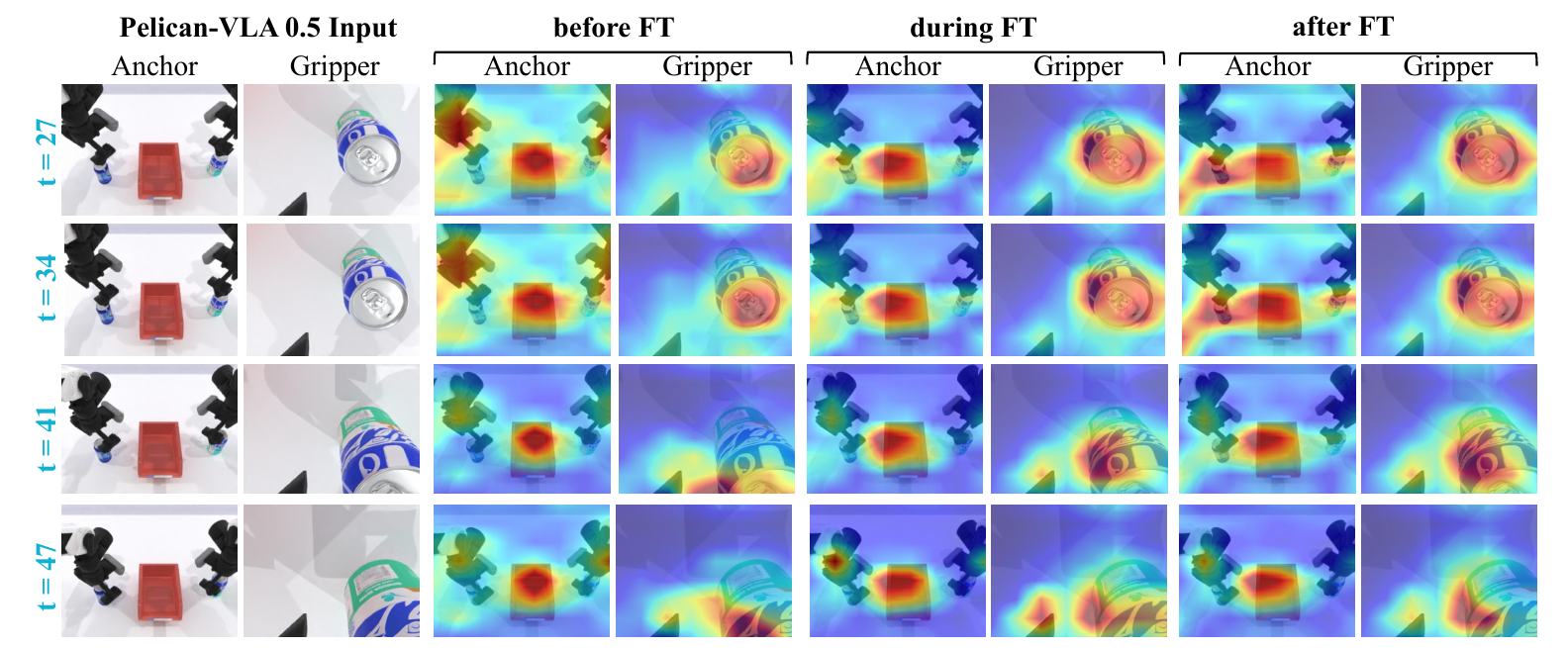}

  \caption{\textbf{Zero-shot and fine-tuned representations are highly similar.} Fine-tuning leaves the BotTokens representation and representation map largely unchanged, yet task success increases substantially. This suggests that the grounded representation is already present before adaptation, while fine-tuning mainly improves the mapping from representation to action.}
  \label{fig:zeroshot_vs_finetune}
\end{figure}

\subsection{Zero-shot vs fine-tuned representations}
\label{sec:analysis:reps}
This report shows that Pelican-VLA 0.5 already develops manipulation-centric attention before any task-specific adaptation. A natural question is whether fine-tuning creates a new representation pattern for the target task, or instead builds on the representation that is already present. To examine this, we compare the zero-shot model and its fine-tuned counterpart on the same held-out task, using attention visualizations, representation maps, and representation-similarity measures.

As shown in Fig.~\ref{fig:zeroshot_vs_finetune}, fine-tuning changes the representation pattern only marginally. The zero-shot and fine-tuned models attend to highly similar object regions, and their BotTokens representations remain close under representation-similarity analysis. Their behavior, however, differs substantially. The zero-shot model often selects and approaches the correct object but fails at fine-grained control, whereas the fine-tuned model executes the task reliably.

\begin{table}[t]
\centering
\caption{Preservation of manipulation-centric attentions during fine-tuning. We compare target attention alignment and attention similarity with respect to the zero-shot model.}
\label{tab:ft_rep_similarity}
\begin{tabular}{lccc}
\hline
\textbf{Stage} &
\textbf{Target Attention IoU} &
\textbf{Attention Similarity}  \\
\hline
Before FT & 0.124 & 1.000 \\
During FT & 0.134 & 0.932 \\
After FT & 0.127 & 0.928  \\
\hline
\end{tabular}
\end{table}

To quantify this observation, we measure target attention IoU, attention similarity, and BotTokens representation similarity between the zero-shot model and fine-tuned checkpoints, as summarized in Table~\ref{tab:ft_rep_similarity}. For the similarity metrics, we compute cosine similarity between temporally aligned attention maps and BotTokens representations from the zero-shot and fine-tuned models. The similarity score measures the consistency of the underlying representation, where a higher value indicates that fine-tuning preserves the corresponding attention pattern or latent representation. The target attention IoU remains stable throughout fine-tuning, while the attention similarity to the zero-shot model remains above 0.9, indicating that task adaptation preserves the manipulation-relevant attention pattern and does not substantially alter where the model attends. Meanwhile, the BotTokens representations undergo moderate adaptation while remaining substantially aligned with the zero-shot representation, suggesting that fine-tuning refines the existing manipulation-centric attention to better support action generation rather than replacing it from scratch.

This result supports the representation-to-action gap introduced earlier. The manipulation-centric attention needed to identify \emph{what} to act upon is largely present before fine-tuning. Fine-tuning does not primarily create this grounding from scratch; rather, it adapts and translates this grounded representation into executable actions.

\section{Conclusion}
\label{sec:conclusion}
This report presents \textbf{Pelican-VLA 0.5}, a unified VLA model built around a compact set of learnable \emph{BotTokens}. Our main finding is that routing perception through this BotTokens-mediated interface induces manipulation-centric attention before task-specific adaptation. Without object labels, segmentation masks, attention supervision, or reasoning traces, the action pathway learns to concentrate on the manipulation-relevant object and its contact region. Controlled comparisons and ablations further show that this behavior is not explained by training data alone. Instead, it is primarily shaped by the BotTokens-mediated route from perception to action, while the BotTokens-language objective makes the representation more controllable by language.

At the same time, Pelican-VLA 0.5 makes clear that zero-shot representation is not the same as zero-shot manipulation. In unseen scenes and unseen robot embodiments, the pre-trained model often identifies and approaches the correct object, but its strict zero-shot success rate remains low. After fine-tuning on RoboTwin, the model achieves state-of-the-art average success among open-source VLA models on this benchmark, while its attention maps and BotTokens representations remain highly similar to those of the zero-shot checkpoint. This suggests that fine-tuning does not primarily create representation from scratch. Rather, it strengthens the mapping from an already-formed manipulation-centric attention to executable actions.

We therefore view Pelican-VLA 0.5 as an intermediate stage between visual representation and practical zero-shot manipulation. The model has begun to answer the question of \emph{what} to act upon, but has not yet fully solved \emph{how} to act upon it across new scenes and embodiments. We refer to this remaining challenge as the \emph{representation-to-action gap}. It is most visible in fine-grained manipulation stages such as stable grasping, contact timing, and precise placement.

We attribute this gap mainly to data scale and action representation. Pelican-VLA 0.5 is trained on only about $2400$ hours of heterogeneous manipulation data, and the data itself remains partially incomplete and noisy. Moreover, the model uses joint-position actions, which are more embodiment-specific than end-effector pose representations and therefore less favorable for cross-embodiment transfer. These limitations likely prevent the grounded representation from being fully converted into reliable zero-shot action competence.

Our next step is to scale both data and action experience. We plan to release a stronger version trained on approximately $7000$ hours of manipulation data, superseding the current $2400$-hour checkpoint. Beyond this, we will continue curating large-scale manipulation data and improving action parameterization, with the goal of turning manipulation-centric attention into robust zero-shot control. To support reproducibility and further research, we will release the code, model weights, and attention-visualization tools.

\pagebreak

\bibliographystyle{abbrv}
\bibliography{biblio}

\pagebreak
\section{Contributions}\label{contributions}
Our contributors are organized based on their roles and magnitude of contribution.
\subsection{Core Contributors}
Data, model, code and pre-training: Zeyuan Ding \\
Code, real-world and simulation experiments: Wenhai Liu\\
Robot deployment and attention visualization: Yang Xu, Jiayu Hu 

\subsection{Contributors}
Yinda Chen, Yi Zhang
\subsection{Tech Lead}
Yong Dai, Zeyuan Ding
\subsection{Corresponding Authors}
Jian Tang, Xiaozhu Ju

\end{document}